%% file: main.tex
\documentclass{article}

\usepackage[final,nonatbib]{nips_2017}

\usepackage[utf8]{inputenc} % allow utf-8 input
\usepackage[T1]{fontenc}    % use 8-bit T1 fonts
\usepackage{hyperref}       % hyperlinks
\usepackage{url}            % simple URL typesetting
\usepackage{booktabs}       % professional-quality tables
\usepackage{amsfonts}       % blackboard math symbols
\usepackage{nicefrac}       % compact symbols for 1/2, etc.
\usepackage{microtype}      % microtypography
\usepackage{graphicx}
\usepackage{amssymb}
\usepackage{epstopdf}
\usepackage{mathtools}
\usepackage{amsmath}
\usepackage[numbers]{natbib}
\usepackage{ntheorem}

\title{An Interpretable and Sparse Neural Network Model for Nonlinear Granger Causality Discovery}

\author{
  Alex Tank, Ian C. Covert, Nicholas J. Foti, Ali Shojaie, Emily B. Fox \\
  University of Washington\\
  \texttt{alextank@uw.edu, icovert@cs.washington.edu, nfoti@uw.edu} \\
  \texttt{ashojaie@uw.edu, ebfox@uw.edu}
}

\begin{document}

\maketitle

\begin{abstract}
While most classical approaches to Granger causality detection repose upon linear time series assumptions, many interactions in neuroscience and economics applications are nonlinear. We develop an approach to nonlinear Granger causality detection using multilayer perceptrons where the input to the network is the past time lags of all series and the output is the future value of a single series. A sufficient condition for Granger non-causality in this setting is that all of the outgoing weights of the input data, the past lags of a series, to the first hidden layer are zero. For estimation, we utilize a group lasso penalty to shrink groups of input weights to zero. We also propose a hierarchical penalty for simultaneous Granger causality and lag estimation. We validate our approach on simulated data from both a sparse linear autoregressive model and the sparse and nonlinear Lorenz-96 model.
  
\end{abstract}

\input{intro_ts}
\input{background_ts}

\input{nng_ts}

\input{sims_ts}
\input{disc_ts}

\bibliographystyle{plain}
\bibliography{nng}

\end{document}

%% file: intro_ts
\section{Introduction}
Granger causality quantifies the extent to which the past activity of one time series is predictive of another time series. When an entire system of time series is studied, networks of interactions may be uncovered. Classically, most methods for estimating Granger causality assume linear time series dynamics and utilize the popular vector autoregressive (VAR) model \cite{Lutekpohl,lozano:2009}. However, in many real world time series the dependence between series is \emph{nonlinear} and using linear models may lead to inconsistent estimation of Granger causal interactions. Common nonlinear approaches to estimating interactions in time series use additive models, where the past of each series may have an additive nonlinear effect \cite{}. However, additive models may miss important nonlinear interactions between predictors so they may also fail to detect important Granger causal connections.

To tackle these challenges we present a framework for interpretable nonlinear Granger causality discovery using regularized neural networks. Neural network models for time series analysis are traditionally used only for prediction and forecasting --- \emph{not} for interpretation. This is due to the fact that the effects of inputs are difficult to quantify exactly due to the tangled web of interacting nodes in the hidden layers. We sidestep this difficulty and instead construct a simple architecture that allows us to precisely select for time series that have no linear or nonlinear effects on the output.
 
We adapt recent work on sparsity inducing penalties for architecture selection in neural networks \cite{} to our Granger causality selection case. In particular, we select for Granger causality by adding a group lasso penalty on the outgoing weights of the inputs, which we refer to as \emph{encoding selection}. We also explore a hierarchical group lasso penalty for automatic lag selection in autoregressive neural network models. When the true network of nonlinear interactions is sparse, this approach will simultaneously select both only a few time series that Granger cause the output series and also the lag of the interactions.  

%% file: background_ts.tex
\section{Background and problem formulation}
Let $x_t \in \mathbb{R}^p$ denote a $p$-dimensional stationary time series. Granger causality in time series analysis is typically studied using the vector autoregressive model (VAR). In this model, the time series $x_t$ is assumed to be a linear combination of the past $K$ lags of the series
\begin{align}
x_t = \sum_{k = 1}^K A^{(k)} x_{t - k} + e_t,
\end{align}
where $A^{(k)}$ is a $p \times p$ matrix that specifies how lag $k$ effects the future evolution of the series and $e_t$ is mean zero noise. In this model time series $j$ does not Granger cause time series $i$ iff $\forall k, A^{(k)}_{ij} = 0$. A Granger causal analysis in a VAR model thus reduces to determining which values in $A^{(k)}$ are zero over all lags. In higher dimensional settings, this may be determined by solving a group lasso regression problem
\begin{align} \label{linear}
\min_{A^{(1)},\ldots, A^{(K)}} \sum_{t = 1}^T \left(x_t - \sum_{k = 1}^K A^{(k)} x_{t - k} \right)^2 + \lambda \sum_{ij} ||(A^{(1)}_{ij}, \ldots, A^{(K)}_{ij})||_2,
\end{align}
where $||.||_2$ denotes the the $L_2$ norm which acts as a group penalty shrinking all values of $(A^{(1)}_{ij}, \ldots, A^{(K)}_{ij})$ to zero together \cite{yuan:2006} and $\lambda > 0$ is a tuning parameter that controls the level of group sparsity.

A \emph{nonlinear} autoregressive model allows $x_t$ to evolve according to more general nonlinear dynamics
\begin{align}
x_t &= g(x_{(t - 1)},\dots, x_{(t - K)}) + e_t \nonumber = \left(g_{1}(x_{(t - 1)}, \ldots, x_{(t - K)}), \ldots, g_p(x_{(t - 1)}, \ldots, x_{(t - K)}) \right)^T + \epsilon_t \nonumber,
\end{align} 

where $g_i$ is a continuous function that specifies how the past $K$ lags influences series $i$. In this context, Granger non-causality  between two series $j$ and $i$ means that the function $g_{i}$ does not depend on the $x_{(t - 1):(t - K)j}$ variables, the past $K$ lags of series $j$. Our goal is to estimate nonlinear Granger causal and non-causal relationships using a penalized optimization approach similar to Problem (\ref{linear}) for linear models.

%\begin{definition} \label{causal_def}
%Time series $j$ is \emph{Granger non-causal} for time series $i$ if for all $x_{(t - 1):(t - K)}$ and each lag $k$ it is the case that $g_{i}(x_{(t - 1)}, \ldots, x_{(t - K)}) = g_{i}(x_{(t-1)}, \ldots, x'_{(t-k)}, \ldots, x_{(t - K)})$ where $x'_{(t - k)} = (x_{(t-k)1}, \ldots, x'_{(t - k)j}, \ldots, x_{(t-k)p})^T$ for all $x'_{(t - k)j}$. 
%\end{definition}

%% file: nng_ts.tex
\section{Neural networks for Granger causality estimation}
We model the nonlinear dynamics with a multilayer perceptron (MLP). In a forecasting setting, it is common to model the full set of outputs $x_t$ using an MLP where the inputs are $x_{(t - 1):(t-K)}$. There are two problems with applying this approach to our case. First, due to sharing of hidden layers, it is difficult to specify necessary conditions on the weights that simultaneously allows series $j$ to influence series $i$ but not influence series $i'$ for $i \neq i'$. Necessary conditions for Granger causality are needed because we wish to add selection penalties during estimation time. Second, a joint MLP requires all $g_i$ functions to depend on the same lags, however in practice each $g_i$ may have different lag orders.

%As an example, one approach would be to consider how Granger non-causality may be given by zero weights in the network:

%\begin{proposition} \label{theorem_mlp}
%Time series $i$ is Granger non-causal for time series $j$ if there exits at least one zero weight in every path of weights across layers from inputs $x_{(t-1):(t-K) i}$ to outputs $x_{tj}$.
%\end{proposition}

%%%something about with no zero weights 

%While intuitive, this result is still hard to enforce in full generality during estimation time because there are an exponential number of possible paths between an input and output in a neural network. We stress that conditions for GC in a neural network are needed because we wish to add penalties that select for Granger causality during estimation time, rather than perform a post hoc analysis of the network after training \cite{something}. 

\subsection{Granger causality selection on encoding}
 To tackle these challenges we model each $g_i$ with a separate MLP, so that effects from inputs to outputs are easier to disentangle. Assume that for each $i$, $g_i$ takes the form of an MLP with $L$ layers and let the vector $h^{l}_{t}$ denote the values of the $l$th hidden layer at time $t$.  Let ${\bf W} = \{W^{1}, \ldots, W^{L}\}$ denote the weights at each layer and let the first layer weights be written as $W^{1} = \{W^{11}, \ldots, W^{1K}\}$ . The first hidden values at time $t$ are given by:
\begin{align} \label{mlp}
h^{l}_{t} = \sigma\left(\sum_{k = 1}^K W^{1k} x_{t - k} + b^{l}\right) 
\end{align} 
where $\sigma$ is an activation function and $b^{l}$ is the bias at layer $l$. The output, $x_{ti}$, is given by
\begin{align}
x_{ti} = w_{O}^{T} h^{L}_{t} + \epsilon_t.
\end{align}
where $w_{O}^{T}$ is the linear output decoder. In Equation (\ref{mlp}), if the $j$th column of the weight matrix, $W^{1k}_{:j}$, is zero for all $k$, then time series $j$ does not Granger cause series $i$. Thus, analogous to the VAR case, one may select for Granger causality by adding a group lasso penalty on the columns of the $W^{k1}$ matrices to the least squares MLP optimization problem for each $g_i$,
\begin{align} \label{gp_mlp}
\min_{{\bf W}}\sum_{t = 1}^{T} \left(x_{it} - g_i(x_{(t-1):(t - K)}, {\bf W} \right))^2 + \lambda \sum_{j = 1}^p ||(W^{11}_{:j}, \ldots, W^{1K}_{:j})||_F.
\end{align}
For large enough $\lambda$, the solutions to Eq. (\ref{gp_mlp}) will lead to many zero columns in each $W^{1k}$ matrix, implying only a small number of estimated Granger causal connections. 

The zero outgoing weights are sufficient but not necessary to represent Granger non-causality. Indeed, series $i$ could be Granger non-causal of series $j$ through a complex configuration of the weights that exactly cancel each other. However, since we wish to \emph{interpret} the outgoing weights of the inputs as a measure of dependence, it is important that these weights reflect the true relationship between inputs and outputs.  Our penalization scheme acts as a prior that biases the network to represent Granger non-causal relationships with zeros in the outgoing weights of the inputs, rather than through other configurations. Our simulation results in Section \ref{simulations} validate this intuition.

\begin{figure}
\centering
\includegraphics[width = .6\textwidth]{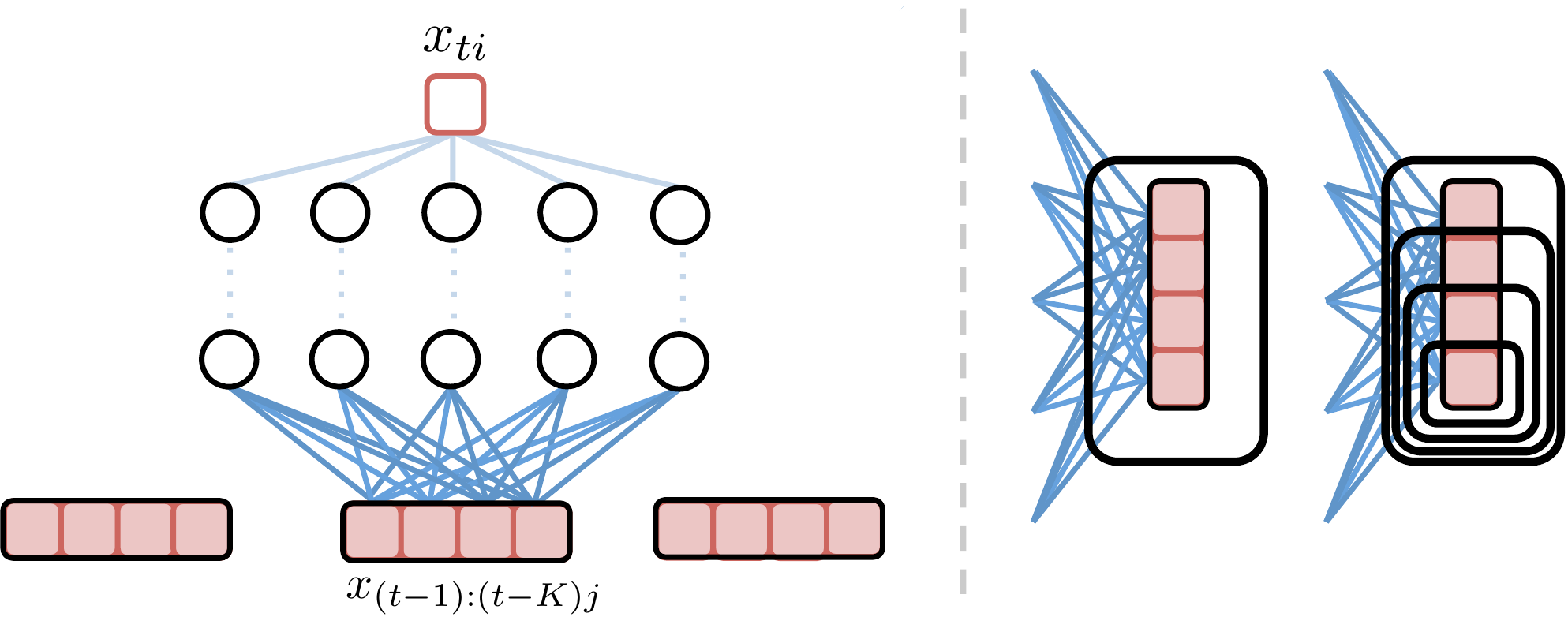}
\vspace{-.1in}
\caption{Schematic for modeling Granger causality using MLPs. (left) If the outgoing weights for series $j$, shown in dark blue, are penalized to zero, then series $j$ does not influence series $i$. (right) The group lasso penalty jointly penalizes the full set of outgoing weights while the hierarchical version penalizes the nested set of outgoing weights, penalizing higher lags more.}
\end{figure}

\subsection{Simultaneous Granger causality and lag selection}
We may simultaneously select for Granger causality and select for the lag order of the interaction by adding a \emph{hierarchical} group lasso penalty \cite{nicholson:2014} to the MLP optimization problem,
\begin{align}\label{hg}
\min_{{\bf W}} \sum_{i = 1}^{T} \left(x_{it} - g_i(x_{(t-1):(t - K)}, {\bf W} \right))^2 + \lambda \sum_{j = 1}^p \sum_{k = 1}^K ||(W^{1k}_{:j}, \ldots, W^{1K}_{:j} )||_F.
\end{align}
The hierarchical penalty leads to solutions such that for each $j$ there exists a lag $k$ such that all $W^{1k'}_{:j} = 0$ for $k' > k$ and all $W^{1k'}_{:j} \neq 0$ for $k' \leq k$. Thus, this penalty effectively selects the lag of each interaction. The hierarchical penalty also sets many columns of $W^{1k}$ to be zero across all $k$, effectively selecting for Granger causality. In practice, the hierarchical penalty allows us to fix $K$ to a large value, ensuring that no Granger causal connections at higher lags are missed. 

%Both Problems (\ref{gp_mlp}) and (\ref{hg}) are optimized using proximal gradient descent. 

%% file: sims_ts.tex
\section{Simulation Experiments} \label{simulations}
\subsection{Linear Vector Autoregressive Model}
\begin{figure} 
\centering
\includegraphics[width=.49\textwidth]{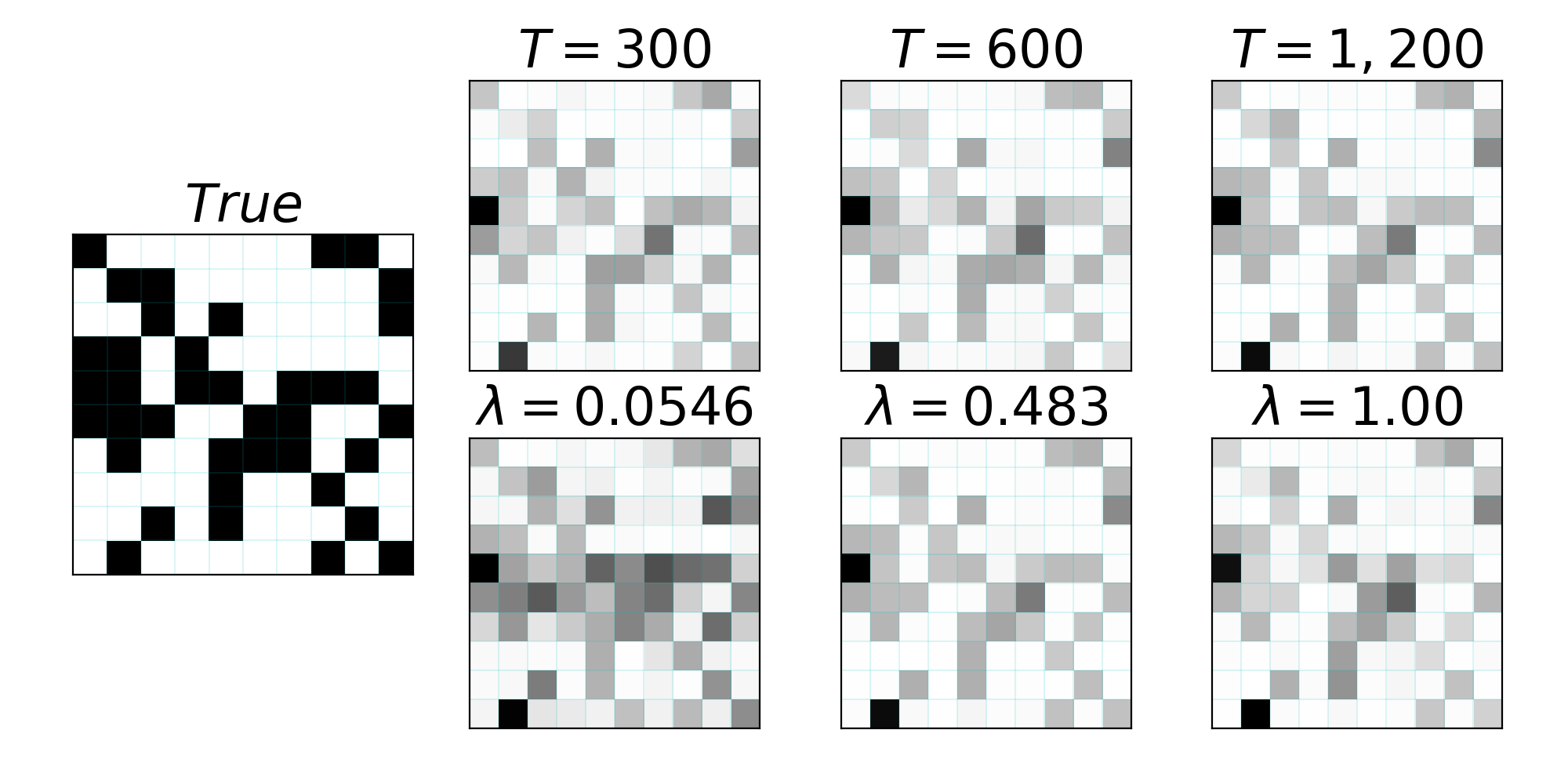}
\includegraphics[width=.49\textwidth]{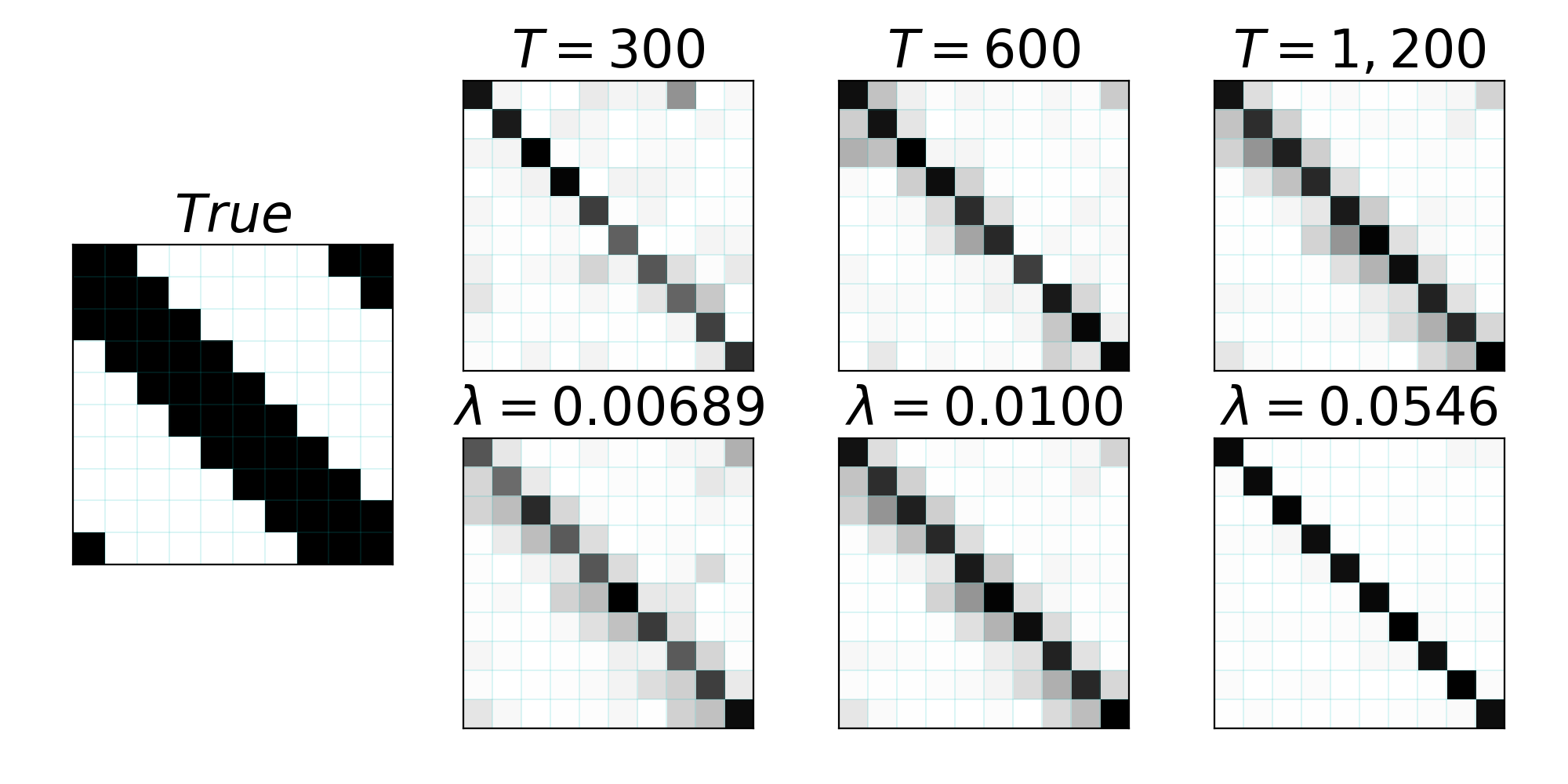}
\vspace{-.2in}
\caption{A comparison of the true graphs to the estimated graphs for multiple $\lambda$ settings and multiple $T$ for data generated from (left) VAR and (right) Lorenz models. Top row in both plots vary $\lambda$ for a fixed $T$ and bottom row vary $T$ with $\lambda$ fixed. The weight of each edge is the $L_2$ norm of the outgoing weights of the respective input series.}
\label{qfig}
\end{figure}

\begin{figure} 
\centering
\includegraphics[width=.3\textwidth]{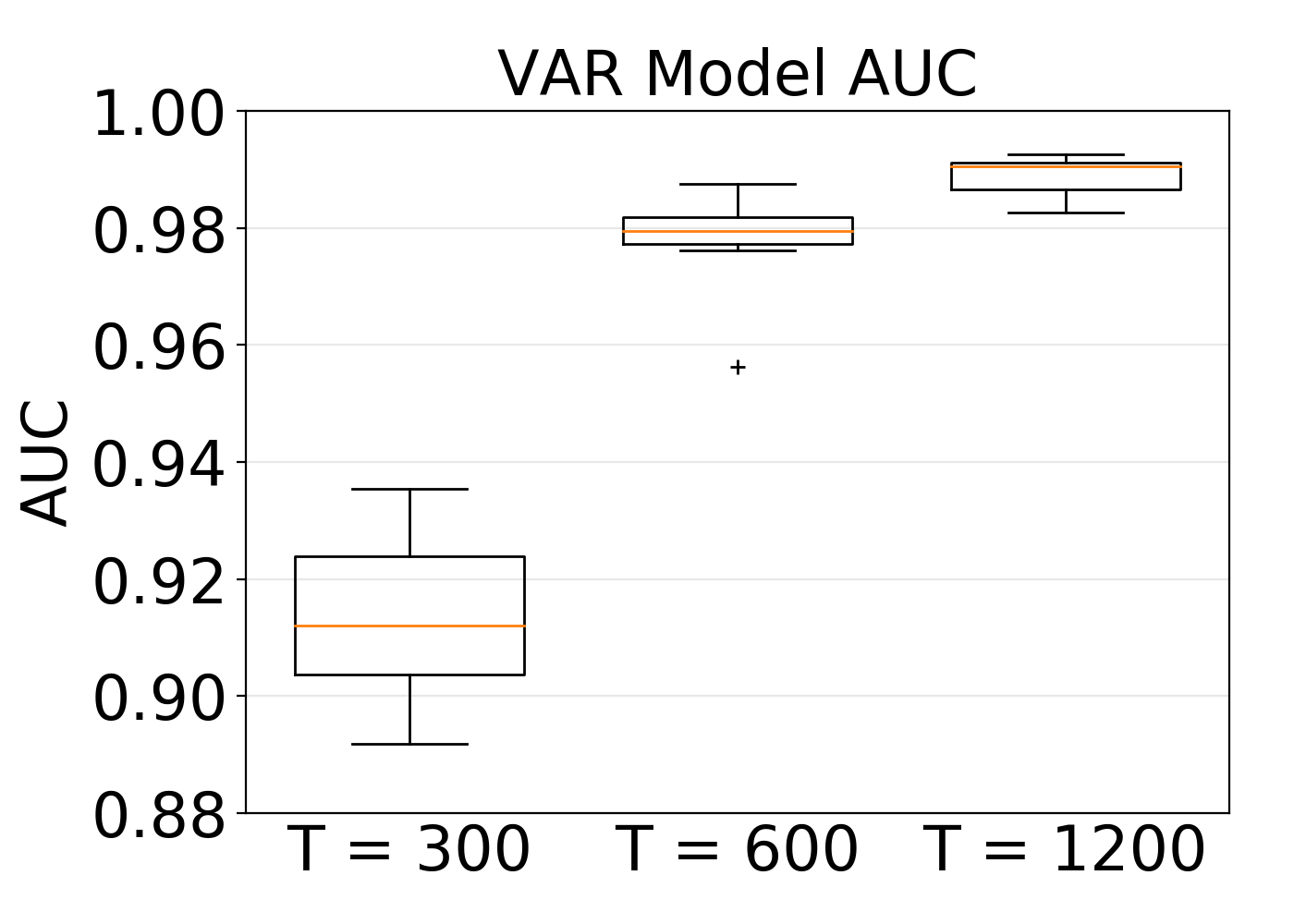} \,\,\,\,\, 
\includegraphics[width=.3\textwidth]{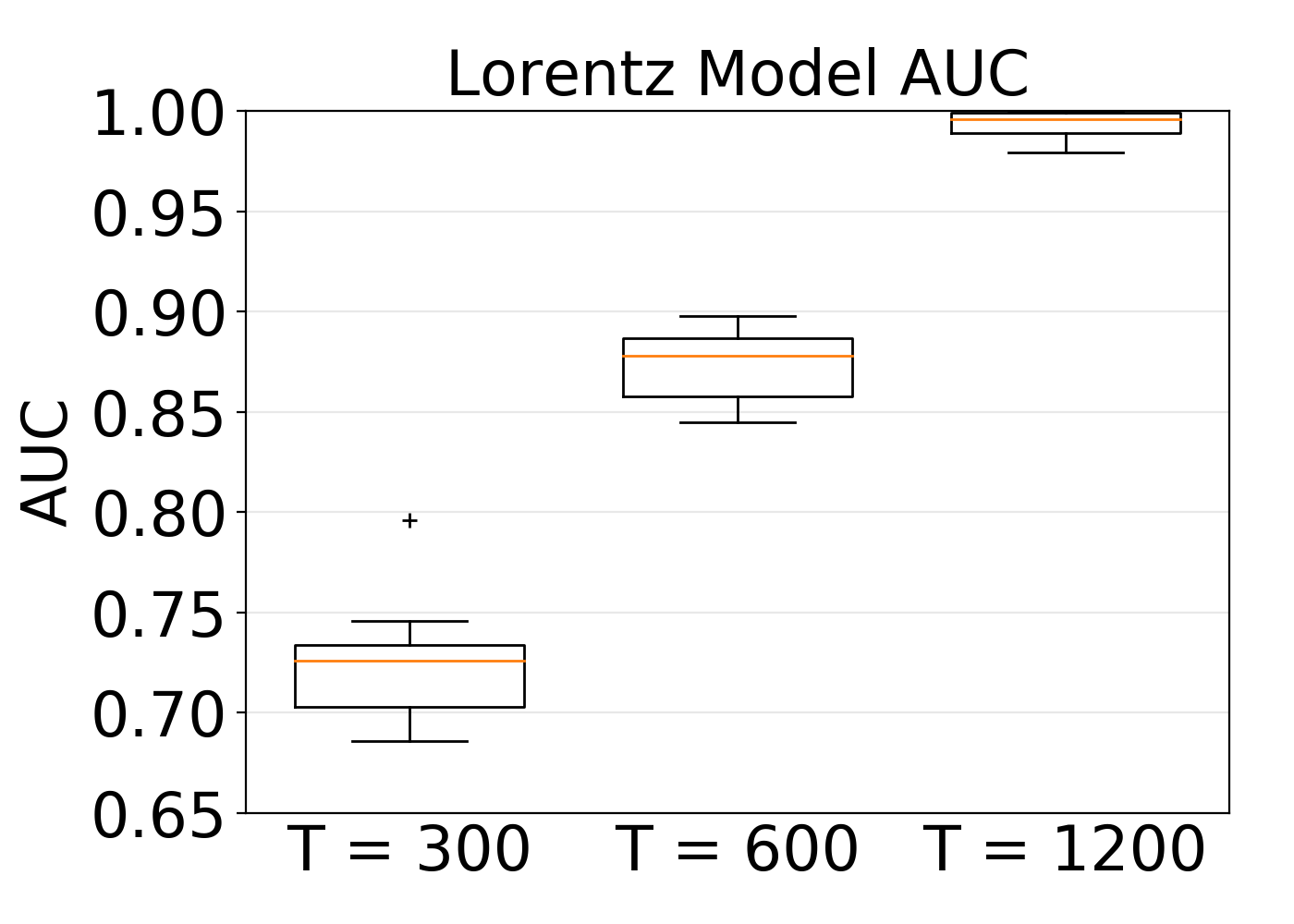}
\vspace{-.0in}
\caption{Box plots of AUC values for data simulated from (left) VAR and (right) Lorenz models.}
\label{aucfig}
\end{figure}

First, we study how our approach performs on data simulated from a VAR model in order to show that it can capture the same structure as existing Granger causality methods. We randomly generate sparse $A$ matrices and apply our group lasso regularization scheme to estimate the Granger causality graph. In Figure~\ref{qfig} (left) we show the estimated graphs for multiple $T$ and $\lambda$ settings and in Figure \ref{aucfig} we show the distribution of AUC values obtained from ROC curves for graph estimation using 10 random seeds. The ROC curves are computed by sweeping over a grid of $\lambda$ values. The AUC values quickly approach the value one as $T$ increases, suggesting that our method is consistent for VAR data.

\begin{figure}
\centering
\includegraphics[width=.9\textwidth]{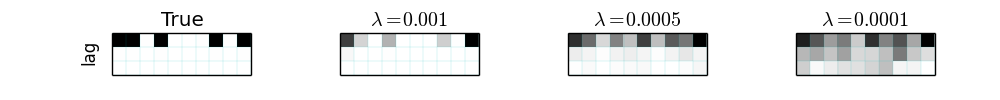}
\vspace{-.2in}
\caption{Estimates of Granger causal interactions for a single series, $x_{ti}$, of a VAR process using the hierarchical penalty. Plots are shown for various $\lambda$ settings.}
\label{hfig}
\end{figure}

To visualize the performance of the hierarchical penalty we show the estimated graph, including lags, for a single $g_i$ on a $p = 10$, $T = 1000$ example in Figure \ref{hfig}. At lower $\lambda$ values both more series are estimated to have Granger causal interactions and higher order lags are included.

\subsection{Nonlinear Lorenz-96 Model}
Second, we apply our approach to simulated data from the Lorenz-96 model \cite{karimi2010}, a nonlinear model of climate dynamics. The dynamics in a $p$-dimensional Lorenz model are
\begin{align} \label{lorenz}
\frac{d x_{ti}}{ d t} = \left(x_{i+1} - x_{i - 2} \right) x_{i-1} - x_{i} + F,
\end{align}
where $x_{-1} = x_{p}$, $x_{0} = x_{p}$, and $x_{N+1} = x_1$ and $F$ is a forcing constant (we take $F=5$). We numerically simulate the Lorenz-96 model using Euler's method, which results in a multivariate, nonlinear autoregressive time series with sparse Granger causal connections. When generating the series with Euler's method, the self connections are much stronger than the off diagonal interactions since the derivative in Eq. (\ref{lorenz}) is multiplied by the Euler step size (we use $.01$).
%For series $i$, only series $i - 2$, $i - 1$, and $i + 1$ have Granger causal interactions. 
We show estimated graphs for multiple $\lambda$ and $T$ values in Figure~\ref{qfig} (right) from data generated from a $p = 10$ model. In Figure~\ref{aucfig} we show box plots of AUC values for multiple $T$ values over 10 random seeds. Overall, the AUC values approach the value one, suggesting consistency of our approach in this nonlinear model.

%\begin{figure} \label{lorfig}
%\centering
%\includegraphics[width=.5\textwidth]{lorentz_comparisons.png}
%\caption{A comparison of the true interaction graph with lags to the estimated graph for a single $x_{ti}$ using the hierarchical group lasso penalty. Plots shown for various $\lambda$ settings.}
%\end{figure}

%% file: disc_ts.tex
\section{Concurrent and future work}
We are currently extending the work in two directions. First, the method herein performs selection on the encoding stage. Alternatively, we could use separate networks to learn features of each series then perform selection on the decoding layer. Second, we are actively working on extensions using recurrent networks based on both long-short term memory  networks \cite{graves:2012} and echo state networks \cite{jaeger:2001}. 

%An alternative approach applicable to multi-output MLPs would be to directly penalize the Jacobian matrix of how the input influences the outputs. In the linear case, this simply reduces to penalizing the linear coefficients. Overall, the work we present herein is the first piece in a series of exciting projects for using penalized deep models for probing nonlinear Granger causal connectivity. 

\paragraph{Acknowledgments}
AT and EF acknowledge the support of ONR Grant N00014-15-1-2380,
NSF CAREER Award IIS-1350133, and AFOSR Grant FA9550-16-1-0038. AS acknowledges the
support from NSF grants DMS-1161565 and DMS-1561814 and NIH grants 1K01HL124050-01 and
1R01GM114029-01.

%% file: main.bbl
\begin{thebibliography}{10}

\bibitem{alvarez:2016}
Jose~M Alvarez and Mathieu Salzmann.
\newblock Learning the number of neurons in deep networks.
\newblock In {\em Advances in Neural Information Processing Systems}, 2016.

\bibitem{basu:2015}
Sumanta Basu, Ali Shojaie, and George Michailidis.
\newblock Network granger causality with inherent grouping structure.
\newblock {\em The Journal of Machine Learning Research}, 2015.

\bibitem{graves:2012}
Alex Graves.
\newblock Supervised sequence labelling.
\newblock In {\em Supervised Sequence Labelling with Recurrent Neural
  Networks}. Springer, 2012.

\bibitem{hastie:1990}
Trevor Hastie and Robert Tibshirani.
\newblock {\em Generalized additive models}.
\newblock Wiley Online Library, 1990.

\bibitem{jaeger:2001}
Herbert Jaeger.
\newblock {\em Short term memory in echo state networks}, volume~5.
\newblock GMD-Forschungszentrum Informationstechnik, 2001.

\bibitem{karimi2010}
A~Karimi and Mark~R Paul.
\newblock Extensive chaos in the lorenz-96 model.
\newblock {\em Chaos: An Interdisciplinary Journal of Nonlinear Science}, 2010.

\bibitem{Louizos:2017}
C.~{Louizos}, K.~{Ullrich}, and M.~{Welling}.
\newblock {Bayesian Compression for Deep Learning}.
\newblock {\em ArXiv e-prints}, 2017.

\bibitem{lozano:2009}
Aurelie~C Lozano, Naoki Abe, Yan Liu, and Saharon Rosset.
\newblock Grouped graphical granger modeling methods for temporal causal
  modeling.
\newblock In {\em Proceedings of the 15th ACM SIGKDD International Conference
  on Knowledge Discovery and Data Mining}, 2009.

\bibitem{lutkepohl:2005}
Helmut L{\"u}tkepohl.
\newblock {\em New introduction to multiple time series analysis}.
\newblock Springer Science \& Business Media, 2005.

\bibitem{nicholson:2014}
W.~B. {Nicholson}, J.~{Bien}, and D.~S. {Matteson}.
\newblock Hierarchical vector autoregression.
\newblock {\em ArXiv e-prints}, 2014.

\bibitem{sindhwani2012scalable}
Vikas Sindhwani, Ha~Quang Minh, and Aur{\'e}lie~C. Lozano.
\newblock Scalable matrix-valued kernel learning for high-dimensional nonlinear
  multivariate regression and granger causality.
\newblock In {\em Proceedings of the Twenty-Ninth Conference on Uncertainty in
  Artificial Intelligence}, 2013.

\bibitem{terasvirta:2010}
Timo Terasvirta, Dag Tjostheim, Clive~WJ Granger, et~al.
\newblock Modelling nonlinear economic time series.
\newblock {\em OUP Catalogue}, 2010.

\bibitem{tong:2011}
Howell Tong.
\newblock Nonlinear time series analysis.
\newblock In {\em International Encyclopedia of Statistical Science}. Springer,
  2011.

\bibitem{yuan:2006}
Ming Yuan and Yi~Lin.
\newblock Model selection and estimation in regression with grouped variables.
\newblock {\em Journal of the Royal Statistical Society: Series B (Statistical
  Methodology)}, 2006.

\end{thebibliography}
